\documentclass[10pt,twocolumn,letterpaper]{article}

\usepackage{cvpr}              

\usepackage{graphicx}
\usepackage{amsmath}
\usepackage{amssymb}
\usepackage{booktabs}
\usepackage{multirow}

\usepackage[pagebackref,breaklinks,colorlinks]{hyperref}

\usepackage[capitalize]{cleveref}
\crefname{section}{Sec.}{Secs.}
\Crefname{section}{Section}{Sections}
\Crefname{table}{Table}{Tables}
\crefname{table}{Tab.}{Tabs.}

\def\cvprPaperID{*****} 
\def\confName{CVPR}
\def\confYear{2022}

\usepackage[dvipsnames]{xcolor}

\usepackage{marvosym}  
\usepackage{graphicx}
\usepackage{caption}
\usepackage{stfloats} 
\usepackage{cuted}
\usepackage{multirow}
\usepackage{tabularx}

\usepackage{tocloft}
\usepackage{etoc}

\usepackage{titletoc}

\newcommand{\listappendixname}{List of Appendices}
\newlistof{appendices}{apc}{\listappendixname}
\newlistof{appfigures}{afg}{List of Case Study Figures}

\usepackage[normalem]{ulem}
\useunder{\uline}{\ul}{}

\extrafloats{100}

\usepackage{multicol}
\usepackage{aliascnt}
\usepackage{adjustbox}

\begin{document}

\title{MMRA: A Benchmark for Evaluating Multi-Granularity and Multi-Image Relational Association Capabilities in Large Visual Language Models}

\newcommand*\samethanks[1][\value{footnote}]{\footnotemark[#1]}
\author{
\small
\textbf{Siwei Wu}\textsuperscript{\includegraphics[scale=0.03]{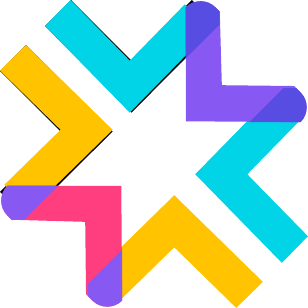},1}\thanks{Eqal authors.}\quad 
\textbf{Kang Zhu}\textsuperscript{\includegraphics[scale=0.03]{figures/map-logo-c.pdf},3}\samethanks[1]\quad 
\textbf{Yu Bai}\textsuperscript{\includegraphics[scale=0.03]{figures/map-logo-c.pdf},3}\samethanks[1]\quad 
\textbf{Yiming Liang}\textsuperscript{\includegraphics[scale=0.03]{figures/map-logo-c.pdf},3}\samethanks[1]\quad 
\textbf{Yizhi Li}\textsuperscript{\includegraphics[scale=0.03]{figures/map-logo-c.pdf},1}\samethanks[1]\quad
\textbf{Haoning Wu}\textsuperscript{\includegraphics[scale=0.03]{figures/map-logo-c.pdf},4}\quad
\\
\small 
    \textbf{J.H. Liu}\textsuperscript{\includegraphics[scale=0.03]{figures/map-logo-c.pdf}}\quad
    \textbf{Ruibo Liu}\textsuperscript{5}\quad
    \textbf{Xingwei Qu}\textsuperscript{\includegraphics[scale=0.03]{figures/map-logo-c.pdf},1}\quad
    \textbf{Xuxin Cheng}\textsuperscript{\includegraphics[scale=0.03]{figures/map-logo-c.pdf},6}\quad 
    \textbf{Ge Zhang}\textsuperscript{\includegraphics[scale=0.03]{figures/map-logo-c.pdf},2,3}\thanks{Corresonding authors.}\quad
    \textbf{Wenhao Huang}\textsuperscript{\includegraphics[scale=0.03]{figures/map-logo-c.pdf},3}\samethanks[2]\quad 
    \textbf{Chenghua Lin}\textsuperscript{0\includegraphics[scale=0.03]{figures/map-logo-c.pdf},1}\samethanks[2]
\\
\small
    \textsuperscript{\includegraphics[scale=0.03]{figures/map-logo-c.pdf}}Multimodal Art Projection Research Community\quad
    \textsuperscript{1}University of Manchester\quad
    \textsuperscript{2}University of Waterloo\quad
    \textsuperscript{3}01.ai\quad
\\
\small
    \textsuperscript{4}National University of Singapore\quad
    \textsuperscript{5}Dartmouth College\quad
    \textsuperscript{6}Peking University\quad
}

\maketitle

\begin{strip}
    \vspace*{-2cm}
    \centering
    \includegraphics[width=\textwidth]{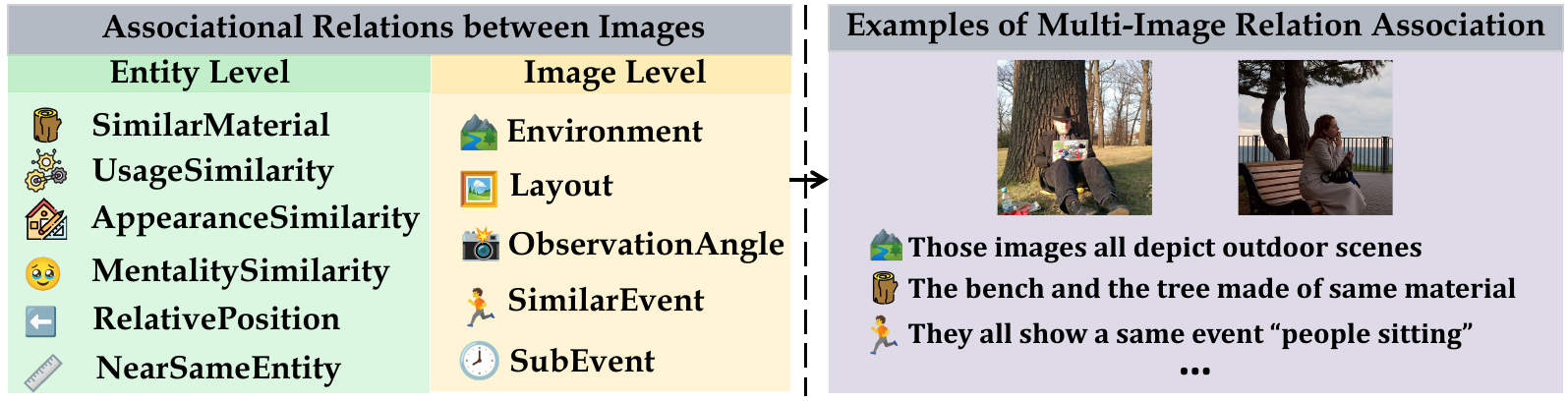}
    \captionof{figure}{Overview of the MMRA benchmark. \textbf{Left}: image Associational Relations extended from the ConceptNet; \textbf{Right}: the examples of Multi-Image Relation Association task.
    }
    \label{fig:framework}
\end{strip}
\begin{abstract}
Given the remarkable success that large visual language models (LVLMs) have achieved in image perception tasks, the endeavor to make LVLMs perceive the world like humans is drawing increasing attention.
Current multi-modal benchmarks primarily focus on facts or specific topic-related knowledge contained within individual images. However, they often overlook the associative relations between multiple images, which require the identification and analysis of similarities among entities or content present in different images.
Therefore, we propose the multi-image relation association task and a meticulously curated \textbf{M}ulti-granularity \textbf{M}ulti-image \textbf{R}elational \textbf{A}ssociation (\textbf{MMRA}) benchmark, comprising \textbf{1,024} samples.
In order to systematically and comprehensively evaluate current LVLMs, we establish an associational relation system among images that contain \textbf{11 subtasks} (e.g, UsageSimilarity, SubEvent, etc.) at two granularity levels (i.e., ``\textbf{image}'' and ``\textbf{entity}'') according to the relations in ConceptNet.
Our experiments reveal that on the MMRA benchmark, current multi-image LVLMs exhibit distinct advantages and disadvantages across various subtasks. Notably, fine-grained, entity-level multi-image perception tasks pose a greater challenge for LVLMs compared to image-level tasks. Moreover, LVLMs perform poorly on spatial-related tasks, indicating that LVLMs still have limited spatial awareness.
Additionally, our findings indicate that while LVLMs demonstrate a strong capability to perceive image details, enhancing their ability to associate information across multiple images hinges on improving the reasoning capabilities of their language model component.
Moreover, we explored the ability of LVLMs to perceive image sequences within the context of our multi-image association task. Our experiments show that the majority of current LVLMs do not adequately model image sequences during the pre-training process.
All our codes and data are released at \url{https://github.com/Wusiwei0410/MMRA}.

\end{abstract}

\section{Introduction}
\label{sec:intro}

Multi-modal perception is a crucial factor for achieving Artificial General Intelligence (AGI) that can perceive the world similarly to humans. 
Significant advancements have been made in multi-modal perception due to the development of Large Visual Language Models (LVLMs)~\cite{li2023blip,liu2024visual,liu2024llavanext,bai2023qwen,ai2024yi}, which have improved visual information processing through diverse visual encoders and bridging the gap between text and image modalities by alignment techniques. 
Consequently, there is growing interest in systematically and comprehensively defining a benchmark to assess the performance of LVLMs and guide future development in this field.
The capabilities of LVLMs in associating multi-image relations can more intuitively and systematically reveal potential shortcomings in VLMs when it comes to multi-image perception tasks. (i.e., if LVLMs struggle to determine the spatial relations within images, they are likely to encounter difficulties in accurately perceiving spatial information when interpreting multiple images).
However, the current multi-modal benchmarks\cite{singh2019towards,MMBench,yue2023mmmu} focus on asking questions within a single image, while the evaluation of LVLMs' multi-image association ability (e.g., ``those images all depict outdoor scenes'' as shown in Fig~\ref{fig:framework}) is overlooked.

There is no comprehensive multi-image benchmark systematically defining the association relations among multiple images to guide the development of multi-image models.
On the one hand, current multi-image benchmarks~\cite{zhao2024benchmarking, DBLP:journals/corr/abs-2406-09411,wu2024scimmir} only focus on monotonous forms of tasks which ask questions about objective facts among pictures like TextVQA~\cite{singh2019towards} (e.g., how many gloves are there in all pictures), or retrieve similar images in a specific domain such as SciMMIR~\cite{wu2024scimmir}. All of these benchmarks overlook the potential connections between images (for example, both images contain objects that are made of wood), which could directly reveal the shortcomings of LVLMs from different perspectives.
On the other hand, mining relations among multiple images at different granularities (i.e., entity level and image level) and across different topics (e.g., the spatial relation and temporal relation between images) present varying degrees of difficulty, and current multi-image benchmarks have not specifically categorized the tasks based on these distinctions.
Although, in the field of Natural Language Processing (NLP), numerous similar works explore the potential relations between textual events or entities~\cite{lin2015sherlock,DBLP:conf/acl/DuDX0022,DBLP:conf/acl/ZhaoCCR23,DBLP:conf/emnlp/GaoHKWMB22,DBLP:conf/naacl/JiangBBC21},
those textual event relations cannot be directly applied to defining the relation among images.

To explore the multi-image perception capabilities of LVLMs, we propose a multi-image relation association task, which requires LVLMs to discern the potential relations between two images (for instance, recognizing that the car and the knife, each present in different images, are both made of iron).
We manually curate a \textbf{M}ulti-granularity \textbf{M}ulti-image \textbf{R}elational \textbf{A}ssociation (\textbf{MMRA}) benchmark,  evaluating the multi-image perception capabilities of LVLMs.
Based on the relations in ConceptNet and observations of potential connections between images, we define an associational relation system, which consists of 6 subtasks at the entity-level granularity (i.e., RelativePosition, NearSameEntity, MentalitySimilarity, AppearanceSimilarity, SimilarMaterial and UsageSimilarity) and 5 subtasks at the image-level granularity (i.e., Layout, Environments, SimilarEvent, SubEvent and ObservationAngle) across from different perspectives of mining relation between images (outline in Fig~\ref{fig:framework}). 
Specifically, we select a subset of the images in LLaVA-665k-multi dataset and employ 5 annotators to manually label 1,024 image pairs with questions and answers on the selected data. 
Besides, to eliminate the answer leakage within the text of the question and its options, we manually remove
 the content in the question and option, which could make LLMs and VLMs directly infer the answer to the question without analyzing the accompanying images.

To explore how the content captured by visual modules in images affects the multi-image perception capabilities of current LVLMs, we employ the LLaVa-Next-110B model to generate detailed descriptions of the images. We then evaluate both LVLMs and LLMs using our MMRA benchmark across four distinct input configurations: Image+Question (IQ), Description+Question (DQ), Image+Description+Question (IDQ), and Question Only (QO). 
We present our key insights as
follows:

\begin{enumerate}
\itemsep -1mm
    \item Based on the results of the IQ and QO setting, we found that closed-source models like GPT-4o, GPT-4v, and Gemini-Flash outperformed all open-source models.  In particular, GPT-4o achieved state-of-the-art (SOTA) performance in the overall evaluation. Additionally, different models exhibited significant performance variations across different subtasks of our benchmark. Some open-source models even surpassed GPT-4 in certain subtasks.
    \item Notably, compared to entity-level tasks, models generally perform better on image-level tasks, and their performance tends to be relatively poor in tasks related to spatial awareness. It indicates that current LVLMs have weak fine-grained multi-image association capabilities and are not proficient in handling spatial perception tasks.
    \item  For image-level tasks, incorporating image descriptions significantly boosts the performance of LLMs, placing them just below GPT-4o and GPT-4v. In contrast, the performance of LVLMs shows no notable improvement with the addition of image descriptions. This indicates that the image-level capability of LVLMs mainly relies on the image content perception ability, and the current LVLMs are limited by the reasoning ability of their language module.
    \item Based on our MMRA benchmark, we also examined the multi-image sequence perception capabilities of LVLMs by altering the order of input image pairs. With the exception of Idefics2, most open-source LVLMs scored relatively low overall, suggesting that they inadequately address the modeling of image sequences during the pre-training phase.
\end{enumerate}

These findings highlight the limitations of current LVLMs in comprehending multi-image scenarios. They underscore the complex perceptual demands at the entity level and reveal that the requirements for visual modules diverge when associating multiple images at both the entity and image levels.

\section{Dataset Curation}

\subsection{Image Pair Selection}
Given that most tasks in the MMRA benchmark require a specific relationship between paired images, we use the semantic similarity of image captions to identify and select image pairs with relatively higher relevance. This aims to reduce the complexity of annotation.
To be specific, we randomly chose the images in the LLaVA-665k-multi to form an image pair. We then utilize the SentenceBERT\cite{reimers2019sentence} to calculate the semantic similarity to filter the image pair with a score below 0.5.
Finally, we obtained 3,403 image pairs for annotation.

\subsection{Subtask Definition}

As shown in the Fig~\ref{fig:benchmark illustration} of Appendix~\ref{appx: Sampled examples from MMRM benchmark}, based on the perspective of humans observing images, we divide our tasks into two granularity levels (i.e., entity and the whole image).
Because the ConceptNet comprehensively defines the relations among different textual event and entity, most of our subtasks, both at entity level and image level, are extended from the relations in ConceptNet.
Additionally, we have also designed some subtasks from a visual perspective (i.e., Layout and ObservationAngle).


\paragraph{Entity level.} We primarily consider the mental state, appearance, and location information of different objects in the images, as well as the psychological characteristics of individual creatures.

\begin{itemize}
\itemsep -1mm
    \item \textbf{RelativePosition (RP)}: The 'AtLocation' is an important relation in ConceptNet to express A is the inherent location of B. As for the entity in two images, we extend this relation into the subtask which judges the relative position of entities in the image. For example, we would ask LVLMs to judge which two entities, respectively in different images, have the same relative position (e.g., all at the upper left of images).
    \item \textbf{NearSameEntity (NSE)}: The relation 'LocatedNear' in ConceptNet expresses ``A and B are typically found near each other''.  Based on it, we design a subtask, 'NearSameEntity', which requires LVLMs to determine whether there are entities, respectively in different images, near the same object.
    \item \textbf{MentalitySimilarity (MS)}: 'HasProperty' in ConceptNet is a relation that describes the characteristics of an entity. We think the emotional property expressed by the images could directly affect humans. Thus, we extend this relation to a subtask that requires LVLMs to determine whether the creatures in two images have similar emotions, attitudes, or feelings (e.g., happy, excited, serious, surprised, etc.).
    \item \textbf{AppearanceSimilarity (AS)}: The physical characteristics of the entity are also an important factor. So we design a subtask that is also relevant to 'HasProperty' and that requires LVLMs to determine whether two images have entities that are physically similar in appearance (e.g., the shape and color of objects, the body and hairstyle of humans).
    \item \textbf{SimilarMaterial (SM)}: The relation 'MadeOf' in ConceptNet expresses 'A is made of B'. Therefore, we design the subtask 'SimilarMaterial' which requires LVLMs to judge whether there are entities, respectively in different images, with the same production materials.
    \item \textbf{UsageSimilarity (US)}: Apart from the aforementioned aspects, we have also devised a subtask that requires LVLMs to discern whether the entities, respectively in two images, have the same usage according to the ConceptNet's relation 'UsedFor' which express 'the purpose of A is B'.
    
\end{itemize}

\paragraph{Image level.} We primarily consider the correlation between the events expressed by the whole image as well as the overall spatial structural similarities of different images.

\begin{itemize}
\itemsep -1mm
    \item \textbf{Layout (LO)}: At the image granularity, we regard the layout of the image as a representation of the relation ``AtLocation''. We design a subtask that requires the LVLMs to determine whether there are similarities in layout between images according to the relation 'NearBy'.
    \item \textbf{Environment (Env)}: From the visual perspective, the environment of the image is also an important content that humans tend to notice (e.g., both images depict the streets of a European country with a Gothic architectural style). So, we design a subtask that lets LVLMs judge if the environments in those images are similar according to the relation 'AtLocation'.
    \item \textbf{SimilarEvent (SimE)}: Excepting the 'SubEvent' relation in ConceptNet, the similar event is also a crucial factor. So we devise a subtask to evaluate the LVLMs' capability to find the same event that happened in the given two images.
    \item \textbf{SubEvent (SubE)}: The temporary relation is an important connection between two images. Therefore, we extend the relation 'SubEvent' to a subtask that requires LVLMs to determine whether the two images describe events that occurred at the same scene in two consecutive moments.
    \item \textbf{ObservationAngle (OA)}: In addition to the layout of the images, we create a subtask for the model to determine whether one of the images is a close-up, an inside shot, or a different parallel angle shot of another image for the sake of comprehensively exploring the view perception ability of LVLMs according to the relation 'LocatedNear'.
    
\end{itemize}

\subsection{Data Annotation}

We hired four postgraduate students specializing in multimodal research to annotate data. Each student was assigned 2-3 tasks.

\begin{figure}[tp]
\centering
\includegraphics[width=0.99\columnwidth]{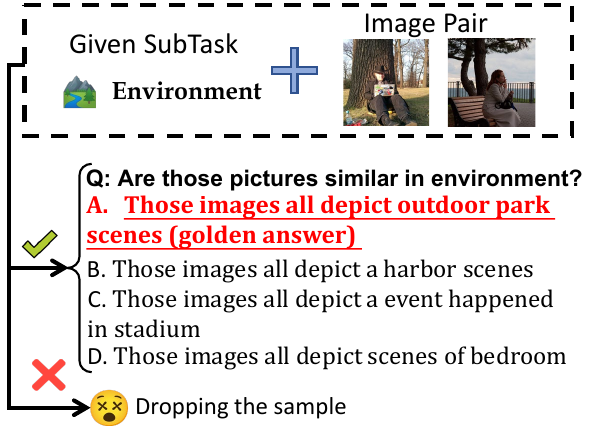}
\caption{The process of annotation.}
\label{figure: Annotation}
\end{figure}

\paragraph{Annotation Process.}
 As shown in Fig~\ref{figure: Annotation}, each annotator is provided with two images and a certain subtask (i.e., Environment). Their responsibility is to determine whether they could design a question based on the given task for the image pair. If the image pair meets the task requirements, they proceed to annotate a question, and options (either multiple-choice or true/false) for that pair. The annotator terminates annotating a task once they reach a predetermined number of labelled samples (i.e., 90) or once all the image pairs for that task have been annotated.

\paragraph{Quality Control.}

We also conducted cross-validation on the annotated data. Specifically, each annotator reviews 2-3 tasks labeled by their peers. If any annotated samples don't meet the task requirements or if the answers derived from the images and options don't match the correct answer, those samples are removed. Additionally, we ensure that the answers to all questions cannot be directly inferred from the text alone (i.e., the question and options). Quality control is concluded once all annotators agree that their verified portion satisfies the specified requirements.

\paragraph{Data Statistics}

\begin{figure}[tp]
\centering
\includegraphics[width=0.85\columnwidth]{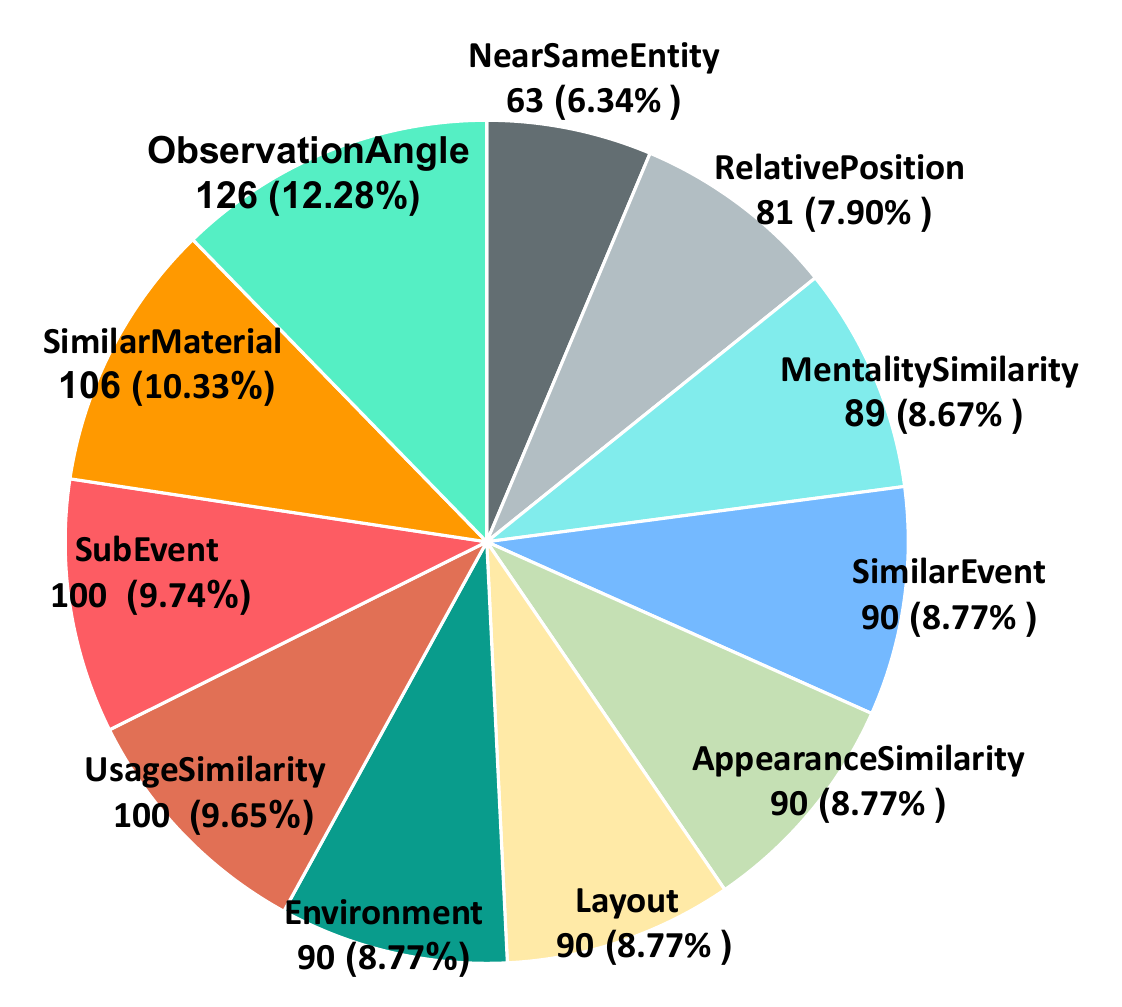}
\caption{The number and ratio of each subtask in our MMRA benchmark. The integers in the graph represent the number of samples in each task, while the percentages in parentheses indicate the proportion of each task in the entire benchmark.}
\label{figure: MMRM Statics}
\end{figure}

As shown in Fig~\ref{figure: MMRM Statics}, we obtain a total of 1,024 annotated samples. To maintain the balance of samples of the subtasks, we endeavored to maintain that the number of samples for all tasks is around 90. The ObservationAngle task has the highest proportion in the entire benchmark, with a total of 126 samples (12.28\%). Due to the difficulty of labeling in the NearSameEntity task, we removed some samples with inconsistent opinions from different annotators during the quality control process and this subtask only has 65 samples.

\subsection{Elimination of Answer Leakage from Questions and Options}

When designing multiple-choice options at the entity level, we need to identify potential entities that could be regarded as the correct answer to the question and provide justifications. For example, as illustrated in Fig~\ref{fig:framework}, 'both tree and bench are made of wood' can be the answer to the SimilarMaterial subtask. However, language models can sometimes deduce the correct answer simply by analyzing the textual content in options. Additionally, annotators often unconsciously label the correct answer with greater detail and specificity, and the language model towards choosing these more detailed options.

To eliminate these biases, we optimize the questions and options for subtasks where the language model scores higher than the expected accuracy by randomly answering the question. For instance, the expected accuracy for true/false questions is 50\%, and for multiple-choice questions with four options, it is 25\%.

\begin{figure}[tp]
\centering
\includegraphics[width=0.99\columnwidth]{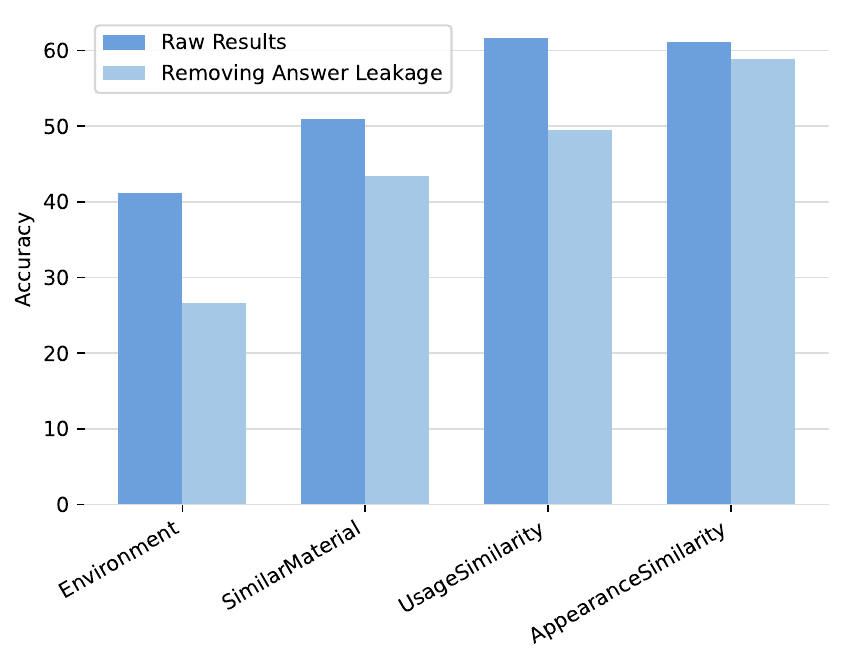}
\caption{Comparing results before and after textual answer leakage elimination.}
\label{figure: Eliminating}
\end{figure}

We refine the options and questions for four subtasks (i.e., UsageSimilarity, Environment, MadeOf, and AppearanceSimilarity), because language models exhibit relatively higher performance on them. As shown in Fig~\ref{figure: Eliminating}, we presented the accuracy changes of the Yi-1.5-9B model before and after answer leakage removal. We have significantly reduced the leakage of answers in the question and option texts. The performances on these subtasks, when using the refined benchmark, are close to the expected random accuracy rates for their respective task types.

For the UsageSimilarity subtask, the performance of language models remains significantly higher than random expectations. We hypothesize that this is because mining the similarity in usage between two entities, a type of general commonsense knowledge, relies heavily on the language model's inference capabilities. Additionally, the current commonsense reasoning capabilities of language models make them adept at identifying subtle differences among the options.

\section{Experiment}
\label{sec:formatting}

\subsection{Experiments Setting}
\label{subsec:Experiments Setting}

To explore the impact of LVLM's image-captioning ability on its multi-image perception, we design four input settings: Image + Question, Description + Question, Image + Description + Question and Question Only. Specifically, we use the LLaVA-NeXT-100B to obtain a detailed textual description of the image in the MMRA benchmark.

\paragraph{Image + Question (IQ).}
In this setting, we just include the image pair and question in the prompt.

\paragraph{Description + Question (DQ).}
To investigate the impact of the image caption capability of LVLMs on the perception of multiple images, we include a detailed description of the image pair and question in the prompt.

\paragraph{Image + Description + Question (IDQ).}

Besides, we also include the image pair, its description, and question in the prompt to compensate for the content of the image that cannot be described in the text.

\paragraph{Question Only (QO).}
For the sake of inspecting whether the answer to the questions in our benchmark is leaked in the textual information of options and questions, we only input the question to let LVLMs answer. 

\subsection{Baselines}

\begin{table*}[!hbtp]
\centering
\scriptsize
\resizebox{1.7\columnwidth}{!}{
\begin{tabular}{l|p{0.5\columnwidth}|r|r}\toprule
\multirow{2}{*}{\textbf{Model}} &\multicolumn{1}{c|}{\multirow{2}{*}{\textbf{Pre-training Data}}} & \multicolumn{1}{c|}{\multirow{2}{*}{\textbf{Supporting Input}}} & \multicolumn{1}{c}{\multirow{2}{*}{\textbf{Parameters}}}  \\
 & & &\\\midrule\midrule
GPT4o\&GPT4v & / & Text, Multi Images, Audio&/   \\\midrule
Gemini-Flash   & / & Text, Multi Images, Audio, Video & /  \\\midrule
Idefics2   & Internet Crawled Data (Wikipedia and OBELICS), Public Multimodal Dataset, LAION-COCO, PDFA (en), IDL, Rendered-text, WebSight & Text, Multi Images &  8B \\ \midrule
Qwen-VL-Chat   & LAION-en, LAION-zh, In-house Data, LAION-COCO, DataComp, Coyo, CC12M, CC3M, SBU, COCO Caption & Text, Multi Images & 8B  \\ \midrule
Phi3v  & / & Text, Multi Images, Video  & 8B   \\ \midrule
Mantis-Idefics2   & Mantis-Instruction dataset & Text, Multi Images & 8B   \\ \midrule
LLaMA-3   & / &Text Only & 8B, 70B   \\ \midrule
Qwen1.5\&Qwen2   & Internet Crawled Data & Text Only & 7B, 32B, 72B   \\\midrule
Yi-Chat\&Yi-1.5-Chat   & Web Documents from Common Crawl &Text Only &9B, 43B   \\
\bottomrule
\end{tabular}
}
\caption{The pre-training information and supporting input of the baselines. "\_" refers to non-public or not fully public data.}\label{tab: model_info}
\end{table*}

As shown in Tab~\ref{tab: model_info}, we evaluated our benchmark on both mainstream closed-source and open-source large models.
Regarding close-source LVLMs, we choose OpenAI's \textbf{GPT4o} and \textbf{GPT4v}, as well as Google's \textbf{Gemini-Flash}.
As for the open-source LVLMs, we mainly evaluate those supporting multi-image inputs (i.e., \textbf{Idefics2 }, \textbf{Qwen-VL-Chat}, \textbf{Phi3v}, \textbf{Mantis-Idefics2}).
Besides, we also assess the open-source LLMs (i.e., \textbf{LLaMA}, \textbf{Qwen1.5-Chat}, \textbf{Qwen2-Chat}) under the text-only input setting.

\subsection{Evaluation Protocol}

\paragraph{Prompt.}
As for each task, we all design a prompt to make LVLMs directly generate textual format answers to the question.
Except including the content described in the section~\ref{subsec:Experiments Setting}, we let LVLMs generate the 'A', 'B', 'C' or 'D' for the choice questions, and 'Yes' or 'No' for the T/F questions.
Besides, we also add the options to the prompt.
As for further details about our prompt design, please refer to the Tab~\ref{tab: prompt} in Appendix~\ref{appx: Designed Template}.

\paragraph{Answer Matching and Metric.}
Because the golden answer in our benchmark is in the format of option id (i.e., 'A', 'B', 'C' and 'D') or judgment (i.e., 'Yes' or 'No'), we design a rule to match the response of LVLMs with the golden answer.
Finally, we use the accuracy of the matching results as the score of those models.
Please refer to the Appendix~\ref{appx: Result Matching Rule} for the details of our designed matching rule.

\section{Result Analysis}
\begin{table*}[hbt!]
\begin{center} 
\footnotesize
\resizebox{2.0\columnwidth}{!}{
    \begin{tabular}{l|l|l|cccccc|ccccc}
\toprule
\multicolumn{1}{l|}{ }                                                                     & \multicolumn{1}{l|}{ } & \multicolumn{1}{c|}{ }                                                                              & \multicolumn{6}{c|}{\textbf{Entity Level}}                                                                           & \multicolumn{5}{c}{\textbf{Image Level}}                                                                            \\
\multirow{-2}{*}{\textbf{Setting}}  & \multirow{-2}{*}{\textbf{Model}}                                                                      & \multirow{-2}{*}{\textbf{Overall}}  & 
RP&	US&	MS&	SM &	AS &	NSE & Env&	LO	&SimE	&SubE	&OA \\

\midrule\midrule
  & GPT4o  & \textbf{67.29} 	& 45.68 	& 66.67 	& 65.17 	& 44.34 	& 68.89 	& \textbf{63.49} 	& \textbf{88.89} & 	47.78 & 	77.78 & 	\textbf{97.00 }& 	70.75

 \\
 & GPT4v  & 66.63  & 	38.75 	 & \textbf{70.71 }	 & 60.67 	 & 44.76  & 	\textbf{71.11}  & 	51.61  & 	87.77  & 	\textbf{64.44}  & 	\textbf{78.89} \textbf{}&  	92.00 	 & 66.04

  \\

& Gemini-Pro & 65.01 &	\textbf{48.15}&	67.68&	69.66	&\textbf{47.17}	&67.78	&56.92	&82.22	&54.44&	60.00&	82.00	&\textbf{73.02}
\\

& Gemini-Flash & 60.33 &	34.56&	66.66&	\textbf{70.78}	&25.47	&68.88	&53.84	&83.33	&60.00&	48.88&	93.00	&57.14
\\

& Idefics2 & 56.93 	& 37.04 	& 65.66 	& 69.66 & 	28.30 	& 44.44 & 	53.97 &	87.78 &	36.67 &	72.22 &	88.00 &	45.24 
 \\
 
 & Mantis-Idefics2  & 57.59 	 & 35.80 	 & 62.63 	 & 68.54 	 & 41.51  & 	52.22 	 & 41.27 	 & 82.22  & 	20.00 	 & 74.44 	 & 91.00 	 & 56.35 
\\

& Phi3v & 51.75 &	48.15 &	64.65 &	62.92 &	47.17 	&61.11 	&46.03 	&86.67& 	34.44 	&56.67 &	51.00 &	20.63 
 \\
\multirow{-8}{*}{\textbf{IQ}} & Qwen-VL-Chat & 47.45 &	37.04 	& 58.59 	& 68.54 &	34.91 &	48.89 &	41.27 &	73.33 	&33.33 	&61.11& 	50.00 &	23.02 
 \\ 
\midrule\midrule
 & LLaMA-3-8B-Instruct	&31.76 &	34.57 	&62.63 &	24.72& 	34.91 	&32.22 &	42.86& 	28.89 &	31.11 &	31.11& 	6.00 &	25.40

\\
& LLaMA-3-70B-Instruct	&23.66 &	38.27 &	60.61 &	12.36 	&26.42 	& \ \ 6.67 	&34.92& 	35.56& 	31.11 &	\ \ 6.67 	&0.00 &	14.29

\\
& Qwen1.5-32B-Chat	&32.36 &	39.51 &	64.65 &	11.24 	&40.57 &	36.67 &	49.21 	&33.33 &	31.11 &	42.22 &	0.00 	&17.46

\\
& Qwen1.5-72B-Chat	&37.11 	&33.33 	&63.64 	&51.69 &	33.96 &	41.11 &	34.92 &	28.89 	&31.11 &	50.00 &	50.00 	&0.00

\\
& Qwen2-7B-Chat	&40.43 	&43.21 &	65.66 &	50.56 &	30.19 	&42.22 &	42.86 &	35.56 	&31.11 	&52.22 &	50.00 &	11.91

\\
& Qwen2-72B-Chat	&38.97 	&35.80& 	64.65 &	46.07 &	45.28 &	46.67 &	39.68 &	27.78 &	31.11 &	48.89 	&44.00 &	\ \ 7.14

\\
& Yi-1.5-9B-Chat	&41.68& 	44.44 &	60.61 &	46.07 &	43.40 &	58.89&	30.16 &	26.67& 	31.11 &	40.00& 	50.00 &	26.98

\\
& Yi-34B-Chat&41.57&	34.57& 	51.52 &	47.19 &	37.74 &	55.56 	&26.98 &	25.56 	&45.56 	&48.89 &49.00 &	32.54

\\
\multirow{-9}{*}{\textbf{QO}} & Yi-1.5-34B-Chat	&26.78 	&25.93 	&63.64 	&39.33& 	43.40&	11.11 &	36.51& 	26.67 	&20.00 	& \ \ 5.56 	& \ \ 7.00 &	17.46 
\\

\bottomrule
\end{tabular}
}
\end{center}
\caption{
The main results of current LVMLs and LLMs on our MMRA benchmark. The IQ and QO represent the Image+Question input and Question Only input, respectively.
}
\label{tab:Main Result}
\end{table*}

\subsection{Overall Analysis}

As shown in Table~\ref{tab:Main Result}, when inputting all question and image pairs (Image+Question setting), the close-source model(i.e., GPT-4v, GPT-4o, Gemini-Pro, and Gemini-Flash) achieves the best performance on our MMRA benchmark, with overall accuracy surpassing 60\%. In contrast, the overall performance of other open-source multi-image LVLMs ranges from 50\% to 60\%, with the exception of Qwen-VL-Chat whose score is only 47.45\%.

Although LVLMs demonstrate varying performances across different subtasks, their average performance at the entity level is generally lower than at the image level. 
The LVLMs' performance is notably high for the Environment and SubEvent subtasks, with most of LVLMs scoring over 80\%. This may be because these subtasks primarily require abstract image-caption information, which the LVLMs have learned during their pre-training phase.

It is worth mentioning that subtasks related to spatial perception (i.e., RelativePosition, NearSameEntity, Layout, and ObservationAngle) remain challenging for LVLMs, as most models achieve accuracy below 50\% for these subtasks.

For the LLMs, both the question and the options are incorporated as inputs to evaluate them on the benchmark (i.e., Question Only setting). The performance of LLMs on the 'UsageSimilarity' task consistently exceeds 60\%, which is comparable to the performance of multi-image LVLMs. This suggests that a significant portion of the reasoning required for the 'UsageSimilarity' subtasks relies on commonsense knowledge inherent in the language model component of LVLMs.

Additionally, at the question-only (QO) setting, we observe that all models, regardless of series or parameter size, demonstrate very similar performance across each task. This consistency indicates that our benchmark effectively mitigates textual information leakage and those subtasks rely on visual information for accurately answering.

\subsection{Impact of Image Input }
As shown in Table~\ref{tab:Main Result}, when provided with both image pairs and questions (i.e., the Image+Question setting), multi-image LVLMs demonstrate significantly better performance compared to LLMs under the QO setting (i.e., Question Only). To highlight the performance improvement of LVLMs due to image input across various tasks, we calculate the average performance of all LLMs on each task as a baseline. By comparing this baseline with the performance of LVLMs, we can quantify the actual enhancement brought about by incorporating images.

\begin{figure*}[!tb]
    \centering
    \includegraphics[width=13cm]{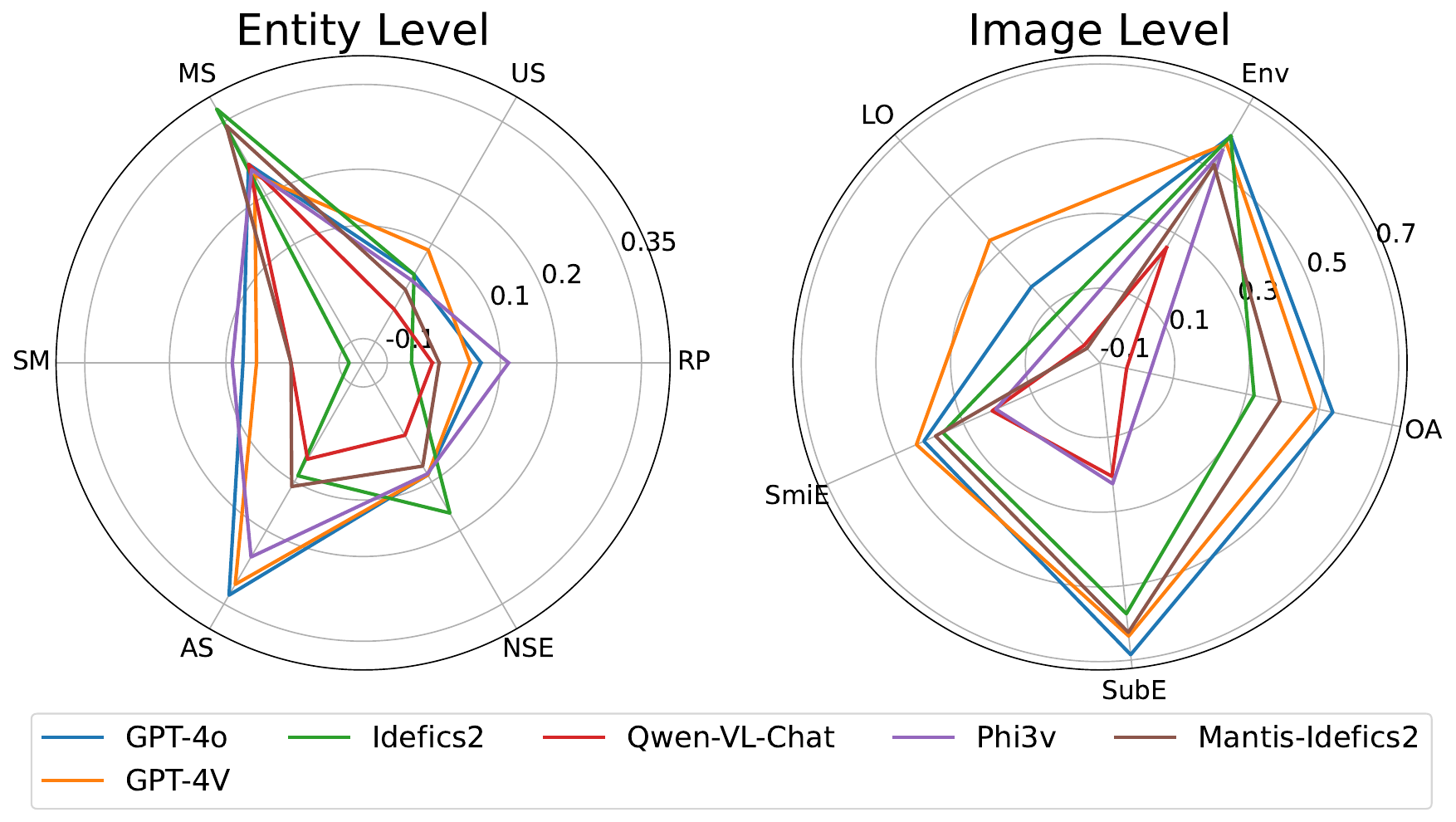}
    \caption{The relative improvement of LVLMs on MMRA benchmark.}
    \label{fig:relative improevment}
\end{figure*}

As shown in the Fig~\ref{fig:relative improevment}, compared to the entity level, the relative improvement at the image level is better, which also indirectly confirms that the entity-level multi-image relation association task requires the model to be able to perceive more image details (the relative improvement at the entity level is around 0.1, while that of the image level is around 0.3).
At the entity level, while the overall performance on the MentalitySimilarity subtask is comparable to other subtasks, the improvement attributed to the inclusion of images is the most significant. 
This suggests that the current LVLMs have developed a robust capacity to perceive mental states during pre-training. As a result, multi-image LVLMs can effectively harness the information in images to analyze the relation between multiple images in the context of individuals' mental states.
This observation offers valuable insight into enhancing current LVLMs. The challenge related to fine-grained multi-image relation association tasks stems from the fact that current multi-image LVLMs do not optimally utilize the detailed information within images. Instead, they tend to rely more on the language model component to process the textual information in the questions.

\subsection{Impact of Image Descriptions}

\begin{table*}[hbt!]
\begin{center} 
\footnotesize
\resizebox{2.0\columnwidth}{!}{
    \begin{tabular}{l|l|l|cccccc|ccccc}
\toprule
\multicolumn{1}{l|}{ }                                                                     & \multicolumn{1}{l|}{ } & \multicolumn{1}{c|}{ }                                                                              & \multicolumn{6}{c|}{\textbf{Entity Level}}                                                                           & \multicolumn{5}{c}{\textbf{Image Level}}                                                                            \\
\multirow{-2}{*}{\textbf{Setting}}  & \multirow{-2}{*}{\textbf{Model}}                                                                      & \multirow{-2}{*}{\textbf{Overall}}  & 
RP&	US&	MS&	SM &	AS &	NSE & Env&	LO	&SimE	&SubE	&OA \\



\midrule\midrule
& LLaMA-3-8B-Instruct	& 54.42 &	46.67 	&60.61 &	46.91 &	77.78 	&29.25& 	57.14 &	47.62 &	57.30 	&57.78 	&62.22& 	51.00 \\
&LLaMA-3-70B-Instruct&	35.86 &	58.89 &	67.68 &	40.74 &	88.89 	&37.74& 	41.27& 	57.14 	&0.00 &	\ \ 5.56 	& \ \ 5.56 &	\ \ 0.00 \\
&Qwen1.5-32B-Chat	&59.52 &	\textbf{66.67} &	67.68 &	40.74 	&86.67 &	37.74 &	53.97 	&43.65 &	59.62 &	67.42 &	73.33 &	52.00 \\
&Qwen1.5-72B-Chat&	60.88& 	51.11 	&\textbf{69.70} &	45.68 &	84.44 &	41.51 	&60.32 	&56.35 &	75.28 &	48.89 &	74.44 &	56.00 \\
&Qwen2-7B-Chat	&54.06& 	32.22 	&64.65 &	39.51 &	85.56 &	32.08 &	60.32 	&30.16 &	58.00 &	61.80 &	48.89 &	\textbf{68.89} \\
\multirow{-6}{*}{\textbf{DQ}}&Qwen2-72B-Chat	&62.17& 	64.44& 	66.67 	&49.38& 	\textbf{92.22} 	&47.17& 	\textbf{63.49}& 	55.56 &	\textbf{69.66} &	50.00 &	72.22 &	51.00 

\\
\midrule\midrule
&Idefics2	&56.35 &	39.51 &	63.64 	&\textbf{75.28}& 	24.53& 	46.67 	&57.14 &	\textbf{88.89} &	33.33 	&68.89 	&\textbf{82.00} &	45.24\\ 
&Qwen-vl-chat	&43.76 &	27.16 &	51.52 &	57.30 	&34.91 &	44.44 &	49.21 &	62.22 	&30.00 &	67.78 &	50.00 &	17.46 \\
&Phi3v	&53.72 &	43.21 &	62.63 &	73.03 &	41.51 &	\textbf{55.56} &	55.56 &	87.78 	&40.00 	&62.22 &	54.00 &	26.98 \\
\multirow{-4}{*}{\textbf{IDQ}}&Mantis-Idefics2&	55.93 &	35.80 &	62.63 &	71.91 	&29.25& 	48.89 &	42.86 &	85.56 &	21.11 	&\textbf{75.56} &	\textbf{82.00} &	55.56 
\\
\bottomrule
\end{tabular}
}
\end{center}
\caption{
The results of DQ and IDQ setting on our MMRA benchmark.
}
\label{tab:DV}
\end{table*}

\paragraph{The key to improving LVLMs' ability to mine associational relations between multiple images lies in enhancing the model's fine-grained perception capabilities.}
We use the LLaVA-NeXT-100B to obtain the image caption and input it as the extra information.
The results of DQ and IDQ settings are presented in the Tab~\ref{tab:DV}.
Under the DQ setting, with the combination of descriptions of image pair, all the LLMs' performance is highly improved, and the overall result of Qwen2-72B-Chat surpasses the Gemini-Flash and is second only to GPT-4v, GPT-4o, and Gemini-Pro.
This also demonstrates that the multi-image understanding capability of LVLMs mainly stems from the content that they precept from images.
As for the DQ setting, after including image descriptions, the performance of LVLMs doesn't change significantly. This suggests that the image descriptions obtained by LLaVA-NeXT-100B overlap with the content perceived by LVLMs themselves. 
Furthermore, it also indicates that the bottleneck in enhancing the multi-image perception capabilities of LVLMs is the reasoning ability of the language model.

\paragraph{Different tasks have varying requirements for the visual or textual components of the LVLMs.}
As for the image level task, the LVLMs' performance is not obviously changed at IDQ setting, while the LLMs' results are close to that of VLMs with the input of images' descriptions.
It demonstrates that the multi-image perception at the image level relies on the visual module of LVLMs.
With regard to the tasks at the entity level, in the IDQ setting, the performance of LVLMs varied the most on the MentalitySimilarity (MS) task, even surpassing GPT-4v and GPT-4o. This indicates that associating the relations of mental states among creatures in multiple images requires highly advanced capabilities from both the language model to process the information obtained from the image.

\subsection{Image Sequence Perception Ability}

Understanding the sequential order of images is crucial for interpreting the relations between multiple images. The ability of a model to comprehend image sequences is essential for tackling complex multi-image tasks, such as sorting images. In certain subtasks of the MMRA Benchmark, the sequence of input images can influence the response to the associated questions.
For instance, in the SimilarMaterial subtask, some options describe entities present in both image1 and image2. Altering the sequence of these input images could lead to a scenario where a correct answer is no longer available.

To examine the ability to perceive image sequences, we adjusted the input image sequence for four specific subtasks: Relative Position, Similar Material, Near Similar Entity, and Observation Angle. Each subtask has options that are directly related to the image sequence. Additionally, we introduced a new option, ``All of the above options are incorrect'' as the correct choice. We then evaluated the performance of LVLMs on these subtasks.

As illustrated in Table~\ref{tab:Sequence}, we present the accuracy metrics of various LVLMs. Idefics2 demonstrates commendable image sequence perception, achieving an overall score close to 60\%. In contrast, most current LVLMs exhibit inadequate image sequence perception abilities, with overall scores below 35\%. This discrepancy suggests that current open-source LVLMs have not adequately addressed image sequence tasks during their pre-training processes.

\begin{table}[tb]
\begin{center} 
\footnotesize
\resizebox{0.9\columnwidth}{!}{
    \begin{tabular}{l|l|cccc}
\toprule
\textbf{Model}& \textbf{Overall}& \textbf{RP}& \textbf{SM}& \textbf{NSE}& \textbf{OA}
 \\
\midrule\midrule
   
Idefics2	&\textbf{59.94}&	\textbf{65.43}&	\textbf{78.30}	&\textbf{82.54}	&\textbf{13.49}\\
Mantis	& \ \ 0.00&	\ \ 0.00&	\ \ 0.00&	\ \ 0.00&	\ \ 0.00\\
Phi3v	&33.20&	41.98&	47.17&	31.75&	11.90\\
Qwen-VL	&\ \ 0.63	&\ \ 0.00&	\ \ 0.94&	\ \ 1.59&	\ \ 0.00 \\

\bottomrule
\end{tabular}
}
\end{center}
\caption{
The results of the Sequence Perception task result.
}
\label{tab:Sequence}
\end{table}

\subsection{Error Analysis}

To better analyze the shortcomings of LVLMs, we examined instances where GPT-4o made errors on relatively challenging subtasks such as RelativePosition, MadeOf, NearSameEntity, and Layout.

\begin{figure*}[!tb]
    \centering
    \includegraphics[width=16cm]{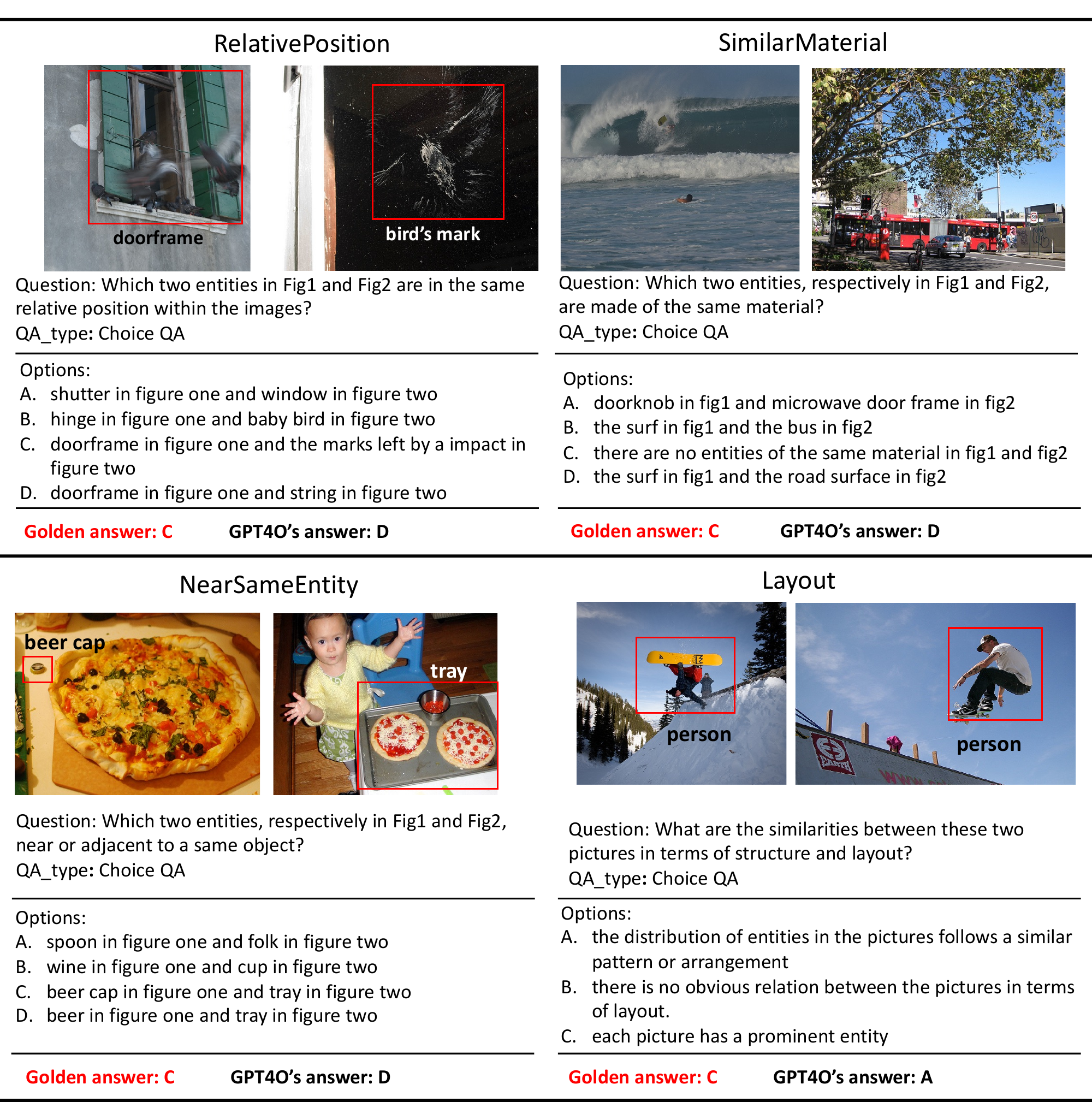}
    \caption{The error analysis of GPT4o on our MMRA benchmark.}
    \label{fig:ErrorAnalysis}
\end{figure*}

As presented in Fig~\ref{fig:ErrorAnalysis}, LVLMs often select entities that do not appear in the image when answering fine-grained questions. For example, for subtasks like 'RelativePosition' and 'NearSameEntity', LVLMs sometimes choose options featuring entities that are not present in the image (e.g., beer and tray).

We believe this issue arises because VLMs primarily depend on the reasoning capabilities of the language model. The textual relations in the options can significantly interfere with the LVLMs' judgments, leading them to overlook the visual input, particularly for fine-detailed questions.

In scenarios where neither image contains the correct answer for the subtask, we introduced an alternative option to express there is no association between two images, such as 'there are no entities of the same material in fig1 and fig2'. When LVLMs cannot identify the correct answer, they tend to select this option, suggesting no connection between the two images.

Regarding the 'Layout' subtask, it appears that current LVLMs have a limited ability to grasp the key elements within images. They sometimes fail to determine whether both images prominently feature a main entity.

\section{Related Work}
\paragraph{Large Visual Language Models.}
With the emergence of Large Language Models(LLMs), researchers have applied it to the multimodal perception field. In recent years, more and more Large Visual Language Models (LVLMs) have achieved excellent success in combining vision knowledge with language on single-image tasks.
BLIP2~\cite{li2023blip} uses a Q-Former to align the visual knowledge with the textual information during the pre-training phase.
LLaVA~\cite{liu2024visual} is pre-trained on the GPT-4 generated multimodal language-image instruction-following data and uses a projection to fuse the vision module and the LM.
LLaVA-Next~\cite{liu2024llavanext} improves the single-image performance with a cost of an increasing number of image tokens.
Based on the framework LLaVA mode, QwenVL~\cite{bai2023qwen}, CogVLM\cite{wang2023cogvlm}, and Yi-VL\cite{ai2024yi} all achieve significant success through a huge number of pre-training data.
Those LVLMs all demonstrate exceptional reasoning ability on single image tasks, such as TextVQA~\cite{singh2019towards}, VQAV2~\cite{goyal2017making}, MMBench\cite{MMBench}, GQA\cite{hudson2019gqa}, etc.
Although Fuyu-8B\footnote{\url{https://www.adept.ai/blog/fuyu-8b}}, Kosmos2~\cite{peng2023kosmos}, and Flamingo~\cite{alayrac2022flamingo} support the interleaved image-text input, they don't optimize in the multi-image reasoning.

\paragraph{Multi-Image Perception Model and Task.}

Currently, some researchers have realized the importance of the multi-image ability of LVLMs. Excepting Kosmos2, Fuyu and Flamingo, There are some models which support input multi images (e.g. Mantis, Idefic2, Phi3v and Mantis-Idefic2~\cite{DBLP:journals/corr/abs-2312-13286, laurenccon2024matters, hanoona2024LLaVA++,jiang2024mantis}).
Besides, the Emu2\cite{DBLP:journals/corr/abs-2312-13286} is the generative multimodal model that supports the interleaved text-image inputs.
And the video understanding models \cite{DBLP:conf/emnlp/ZhangLB23,Ren2023TimeChat} also have the multi-image perception ability and the Video-LLaMA~\cite{DBLP:conf/emnlp/ZhangLB23} also support the multi-image inputs, but it is relatively worse than the VLMs.
Meanwhile, there is also a lack of comprehensive and systematic evaluation of multi-image LVLMs.
The difference description is the first task and researchers have developed many datasets, such as Spot-the-Diff and Birds-to-Words~\cite{DBLP:conf/emnlp/JhamtaniB18}, etc. 
However, they are all generative tasks.
Recently, the MuirBench~\cite{DBLP:journals/corr/abs-2406-09411} and the multi-image under standing benchmark~\cite{zhao2024benchmarking} focus on evaluating the LVLMs' ability, but they don't systematically define the relation among images in real-life scenario.

\paragraph{Commonsense Reasoning.}
During the previous research in NLP, there are numerous works for commonsense reasoning~\cite{DBLP:conf/acl/DuDX0022,DBLP:conf/acl/ZhaoCCR23,DBLP:conf/emnlp/GaoHKWMB22,DBLP:conf/naacl/JiangBBC21} and would use many pre-defined commonsense knowledge (i.e., Knowledge Graph~\cite{DBLP:conf/aaai/SapBABLRRSC19,speer2017conceptnet,DBLP:conf/acl/ShenWX23}).
Such as the Abductive Commonsense Reasoning task using real-world commonsense reasoning to infer the most plausible explanation between two events. Moral Stories~\cite{DBLP:conf/emnlp/EmelinBHFC21} is to generate the next event under different norm restrictions.
In order to help address those commonsense reasoning tasks, researchers define some Knowledge Graph (KG), which preserved knowledge in triplet format (i.e., $<$Head, Relation, Tail$>$).
The ConceptNet~\cite{speer2017conceptnet} and ATOMIC~\cite{DBLP:conf/aaai/SapBABLRRSC19} are Commonsense Knowledge Graphs (CSKGs) where they define numerous relations between event node and entity node.
The current multi-image benchmarks~\cite{DBLP:journals/corr/abs-2406-09411,zhao2024benchmarking} don't define the relation system among images. Although VCD~\cite{shen2024vcd} uses the knowledge system in ConceptNet to mine the potential knowledge in a single image, it cannot directly apply to the multi-image setting.
In this work, we will define a relation system among different images and curate a benchmark.

\section{Conclusion}
The multi-image perception capabilities of LVLMs are often overlooked. To comprehensively and systematically assess these capabilities, we establish a relational system among images based on the knowledge relations in ConceptNet and manually annotate a multi-granularity, multi-image relation association benchmark (MMRA). This benchmark consists of 11 subtasks, divided into image and entity levels. During the curation phase, we carefully consider and mitigate the risk of answer leakage in the questions and options. We refine the benchmark and design experiments to demonstrate that we have largely addressed the issue of answer leakage.
Our evaluation of current multi-image LVLMs on this benchmark reveals significant shortcomings, particularly in fine-grained (entity-level) and spatial perception subtasks. Compared the results of Image+Description+Question setting with Image+Question setting, our experiments highlight that these models lack robust image detail perception capabilities.

{\small
\bibliographystyle{ieee_fullname}
\bibliography{egbib}
}

\clearpage
\appendix

\begin{figure*}[!tb]
    \centering
    \includegraphics[width=14cm]{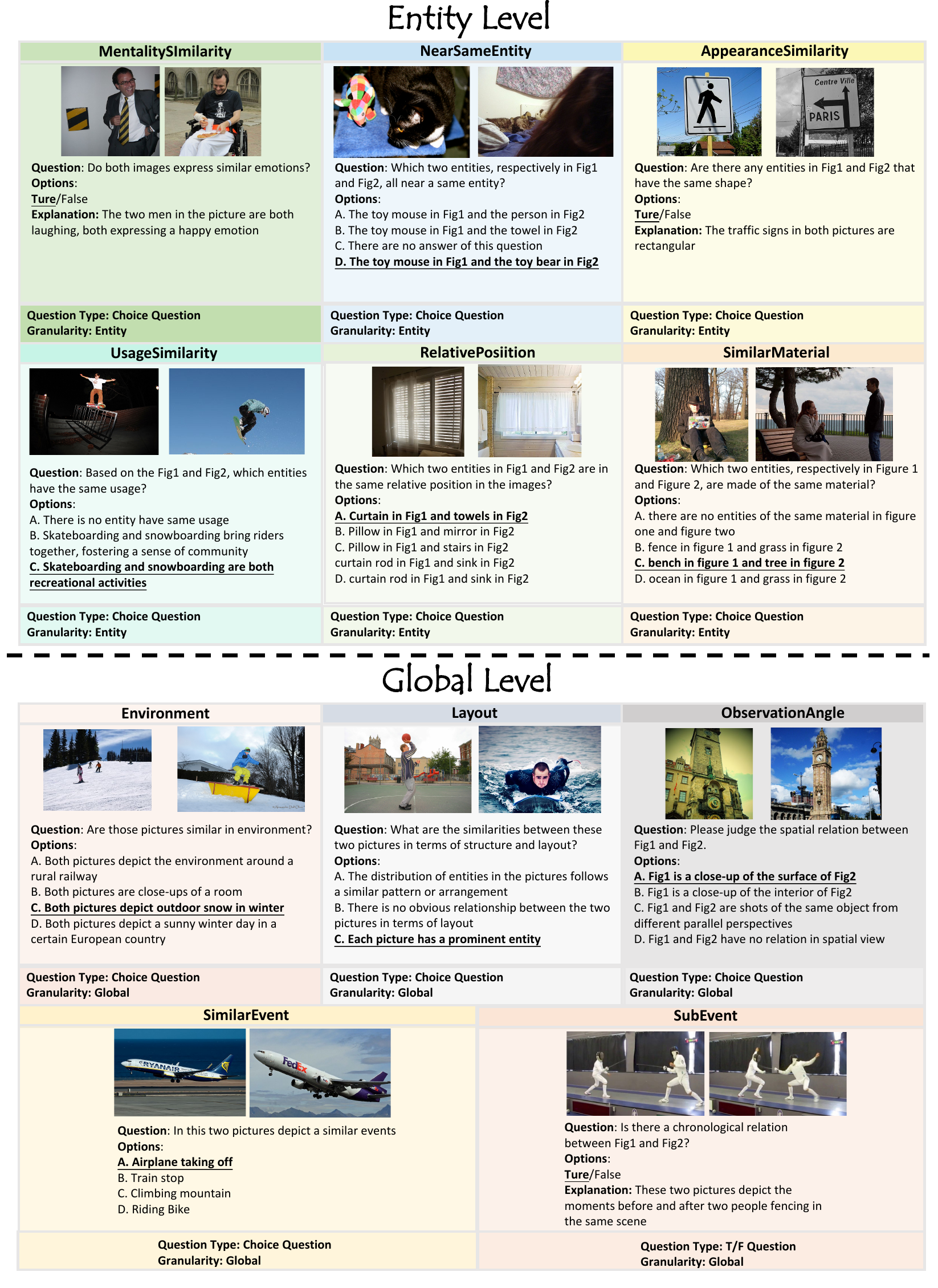}
    \caption{Sampled MMRA examples for each task. The bold and underlined options indicate they are the golden answers.}
    \label{fig:benchmark illustration}
\end{figure*}

\section{Result Exact Matching Rule}
\label{appx: Result Matching Rule}
Due to significant differences in the response styles of various LLMs and chat templates, the content format of model answers can vary greatly. To gap this discrepancy and accurately reflect the responses of different models, we have developed a specialized Exact Matching Rule.

\noindent\textbf{For Multiple-Choice questions:}
First, we use regular expressions to attempt to directly extract the matching content within parentheses, i.e., extracting Answer: ``A'' from ``(A)''. If this is unsuccessful, we then attempt to match option labels (A-D) from the entire response content and return the option with the highest match count. If the response does not contain any option label information, we try to match the option content directly within the response and return the corresponding option label.

\noindent\textbf{For True/False questions:}
We use regular expressions to match ``yes'' or ``no'' within the response content. If there are multiple matches, we return the result that appears the most frequently.

\section{Designed Template}
\label{appx: Designed Template}
In this part, we present our designed prompt template for both Choice Question and T/F Question in the Tab~\ref{tab: prompt}





\begin{table}[tb]
\centering
\scriptsize
\resizebox{0.99\columnwidth}{!}{
\begin{tabular}{l|p{0.5\columnwidth}}\toprule
\multirow{1}{*}{\textbf{Question Type}} &\multicolumn{1}{c}{\multirow{1}{*}{\textbf{Prompt Template}}}   \\ \midrule
T/F Question & You will be giving one question and two images. Please only answer the question with Yes or No. Questions: \{question\}. Please give me your answer.    \\\midrule
Choice Question   &  You will be giving one question, two images,  and four options, one of them is correct. Please choose one of the four options. The question is: \{Question\}.
                    The options are:  [A: \{A\},  B: \{B\},  C: \{C\}, D: \{D\}]
                    Please tell me the answer in the format if [A], [B], [C] or [D]. \\
\bottomrule
\end{tabular}
}
\caption{The designed prompt template for the task in our MMRA benchmark. }\label{tab: prompt}
\end{table}

\section{Sampled examples from MMRA benchmark}
\label{appx: Sampled examples from MMRM benchmark}
In order to comprehensively show our benchmark, we select a sample for each task and present then in the Tab~\ref{fig:benchmark illustration}.
We design two kinds of tasks (i.e., Choice Question and T/F Question). For each example, we show the image pair, question and options.

\end{document}